%% file: CLESS_arxiv_version.tex
\documentclass{article} 
\usepackage[table,xcdraw]{xcolor} 
\usepackage[accepted]{arx2021}
\usepackage{microtype}
\usepackage{graphicx}
\usepackage{subfigure}
\usepackage{booktabs} 

\usepackage{amssymb}
\usepackage[title]{appendix}
\input{math_commands.tex}

\makeatletter
\renewcommand\tagform@[1]{\maketag@@@{\ignorespaces#1\unskip\@@italiccorr}}
\makeatother

\usepackage{longtable}
\usepackage{url}
\usepackage{xargs} 
\usepackage{xparse}  
\usepackage{xspace} 
\usepackage{footmisc}
\usepackage{float}
\usepackage{hyperref}
\usepackage{cleveref}
\usepackage{soul}
\usepackage{graphicx}
\usepackage[outline]{contour}

\usepackage{xr}

\makeatletter
\newcommand*{\addFileDependency}[1]{
  \typeout{(#1)}
  \@addtofilelist{#1}
  \IfFileExists{#1}{}{\typeout{No file #1.}}
}
\makeatother

\usepackage{tikz}
\DeclareRobustCommand*\Circled[1]{\tikz[baseline=(char.base)]{
            \node[shape=circle,draw,inner sep=.14pt] (char) {#1};}} 

\crefname{section}{\S}{\S\S}
\Crefname{section}{\S}{\S\S}
\crefname{table}{Tab.}{}
\crefname{figure}{Fig.}{}
\crefname{algorithm}{Algorithm}{}
\crefname{equation}{eq.}{}
\crefname{appendix}{App.}{}
\crefname{prop}{Proposition}{}
\crefformat{section}{\S#2#1#3}  

\newcommandx{\unsure}[2][1=]{\todo[linecolor=red,backgroundcolor=red!25,bordercolor=red,#1]{#2}}
\newcommandx{\change}[2][1=]{\todo[linecolor=blue,backgroundcolor=blue!25,bordercolor=blue,#1]{#2}}
\newcommandx{\info}[2][1=]{\todo[linecolor=OliveGreen,backgroundcolor=OliveGreen!25,bordercolor=OliveGreen,#1]{#2}}
\newcommandx{\improvement}[2][1=]{\todo[linecolor=yellow,backgroundcolor=yellow!25,bordercolor=yellow,#1]{#2}}
\newcommandx{\thiswillnotshow}[2][1=]{\todo[disable,#1]{#2}}

\usepackage{color}
\definecolor{three}{HTML}{1f77b4}
\definecolor{light}{rgb}{0.28, 0.28, 0.28}
\definecolor{grey}{rgb}{0.35, 0.35, 0.35}
\definecolor{black}{rgb}{0, 0, 0}
\definecolor{revised_blue}{rgb}{0.14, 0.14, 0.63}
\def\light#1{\emph{{\color{black}#1}}}

\newcommand{\pting}{pretraining\xspace}

\begin{document}
\twocolumn[
\arxtitle{Data-Efficient Pretraining via Contrastive Self-Supervision}









\begin{arxauthorlist}
\arxauthor{Nils Rethmeier}{DFKI,CPU}
\arxauthor{Isabelle Augenstein}{CPU}
\end{arxauthorlist}

\arxaffiliation{DFKI}{German Research Center for AI, Germany}
\arxaffiliation{CPU}{Copenhagen University, Denmark}

\arxcorrespondingauthor{}{nils.rethmeier@dfki.de}
\arxcorrespondingauthor{}{augenstein@di.ku.dk}

\vskip 0.3in
]
\printAffiliationsAndNotice{}
\input{main_text}
\bibliography{biblio}
\bibliographystyle{arx2021}
\end{document}

%% file: math_commands.tex
\usepackage{amsmath,amsfonts,bm}

\def\eqref#1{equation~\ref{#1}}

\def\1{\bm{1}}

\def\vzero{{\bm{0}}}
\def\vone{{\bm{1}}}

\def\vl{{\bm{l}}}

\def\vp{{\bm{p}}}

\def\vt{{\bm{t}}}

\def\vw{{\bm{w}}}

\def\mL{{\bm{L}}}
\def\mM{{\bm{M}}}

\DeclareMathAlphabet{\mathsfit}{\encodingdefault}{\sfdefault}{m}{sl}
\SetMathAlphabet{\mathsfit}{bold}{\encodingdefault}{\sfdefault}{bx}{n}

\newcommand{\R}{\mathbb{R}}
\newcommand{\N}{\mathbb{N}}

\newcommand{\sigmoid}{\sigma}



%% file: main_text.tex
\begin{abstract} 
For natural language processing `text-to-text' tasks, the prevailing approaches heavily rely on \pting large self-supervised models on increasingly larger `task-external' data. Transfer learning from high-resource pretraining works well, but research has focused on settings with very large data and compute requirements, while the potential of efficient low-resource learning, without large `task-external' \pting, remains under-explored.
In this work, we evaluate against three core challenges for resource efficient learning. Namely, we analyze: (1) \pting data ($X$) efficiency; (2) zero to few-shot label ($Y$) efficiency; and (3) long-tail generalization, since long-tail preservation has been linked to algorithmic fairness and because data in the tail is limited by definition.
To address these challenges, we propose a data and compute efficient self-supervised, contrastive text encoder, pretrained on 60MB of `task-internal' text data, and compare it to RoBERTa, which was pretrained on 160GB of `task-external' text.
We find our method outperforms RoBERTa, while \pting and fine-tuning in a 1/5th of RoBERTa's fine-tuning time.
\end{abstract}
\section{Introduction}
Self-supervised \pting over \light{large-scale `task-external' data} has been empirically demonstrated as a viable approach to transfer learning in NLP within the framework of `text-to-text' prediction proposed by \cite{T5}. However, the question whether self-supervised `text-to-text' \pting is able to learn effectively from limited \pting data is still not well understood as current models focus on exploring large to Web-scale, `end-task external \pting data' \cite{BLUNSOM,BlunsomTransformerDataInefficient} and have comparably cost-inefficient \pting compute requirements \cite{HARDWARELOTTERY,SEED_HACKING,ENERGY_COST} -- e.g. the now moderately sized RoBERTa was pretrained on 1000 V100 GPUs for one day, on over 160GB of \pting data.

\begin{figure*}[t]
    \centering
    \includegraphics[width=1\linewidth, trim=6 6 6 6, clip]{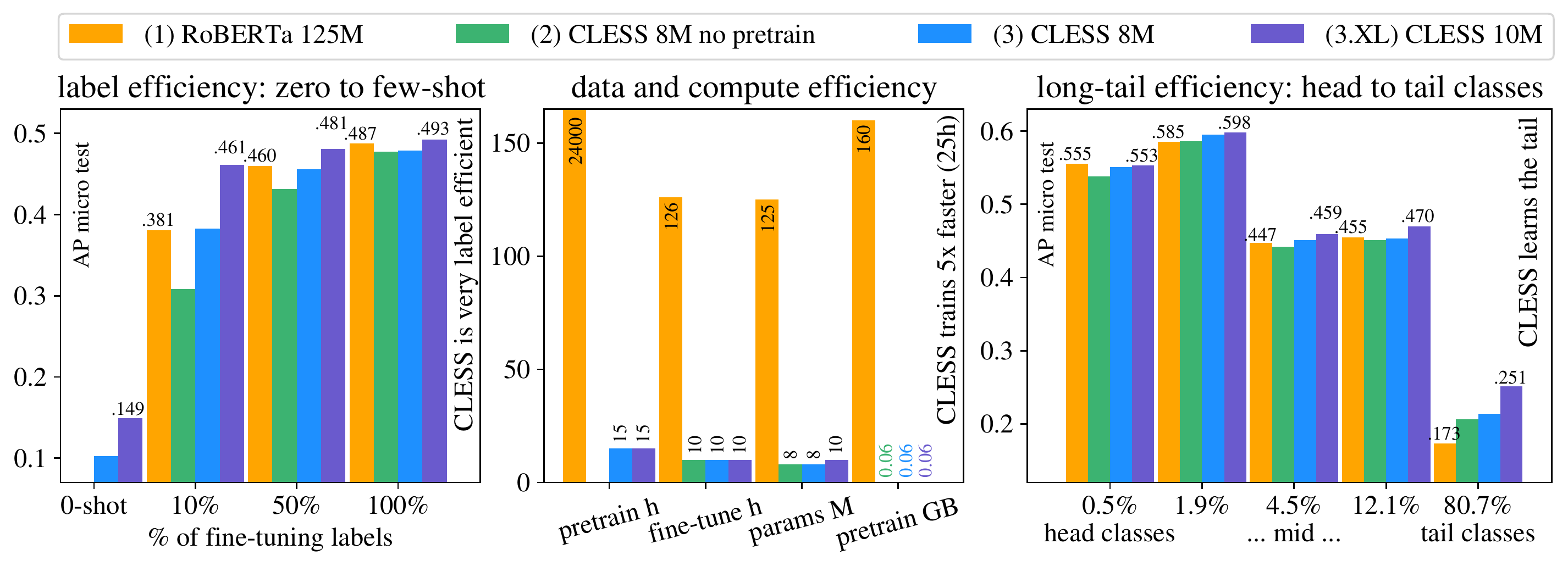}
    \caption{\textbf{CLESS learns labels, data and long-tail generalization more efficiently than RoBERTa:} Models (1), (3) and (3.XL) were pretrained with either a masked language model objective (1) or our contrastive self-supervision objective (3,3.XL) -- details \cref{sec:model}. CLESS 'large' (3.XL) outperforms RoBERTa in zero to few-shot (left plot) and long-tail learning, i.e.\ 80\% of classes (right plot), while (middle plot) pretraining and fine-tuning in 1/5th (15+10 hours) of the time it takes RoBERTa to fine tune only (126 hours). CLESS needs no external pretraining data and pretrains 15 hours, while RoBERTa used 160GB of external texts to pretrain on 1000 V100 GPUs for 1 day.}\label{fig:overview_results}
\end{figure*}

In this work, as seen in \cref{fig:overview_results}, we instead propose to study (data) efficient NLP \pting by using 6-60MB of `task-internal' text data for \pting, and by evaluating learning efficiency aspects including: (1) \pting data efficiency, (2) label data efficiency and (3) long-tail generalization performance. \textit{Notably, our method outperforms a fine-tuned RoBERTa model in all three aspects.}

To study (1) \pting data efficiency (or $X$-efficiency), we propose a novel self-supervised contrastive \pting method, CLESS, short for \underline{C}ontrastive \underline{L}earning-Data \underline{E}fficient \underline{S}elf-\underline{S}upervision.
We pretrain CLESS on 60MB of `task-internal' texts and compare its performance to a fine-tuned RoBERTa -- \cref{sec:few-shot}.
To analyze (2) label fine-tuning efficiency ($Y$-efficiency), we also evaluate RoBERTa and CLESS in few-shot settings using between 10\% and 100\% of labeled instances during supervised end-task learning -- \cref{sec:few-shot}. Further, we analyze the zero-shot prediction performance of CLESS in \cref{sec:zero-shot}, since ``fine-tuning probes introduce uncontrolled external biases into text representation model evaluation, while zero-shot probing avoids these biases'', as \citet{ZERO_SHOT_PROBES,elazar2020amnesic} point out.
Finally, for efficiency aspect (3), we evaluate how well CLESS and RoBERTa generalize to long-tail classes in \cref{sec:longtail}. 

We analyze long-tail learning because training data in the tail is always limited to few or even zero-shot scenarios as seen in \cref{fig:longtail_dist} or pointed out in \cite{LONG_TAIL_PROBLEMS}. Additionally, tail information preservation has been linked to algorithmic bias reduction by \cite{hookerCompressedLongtail,COMPRESSION_KILLED_THE_LONG_TAIL,HookerPatterns}, while current, large-scale \pting typically does not consider reduction approaches to issues of algorithmic bias \cite{TIMNIT_Algorithmic_bias,BIAS,BIG_MODELS}. In \cref{sec:longtail}, we demonstrate that compared to RoBERTa, CLESS pretrained with small data and much fewer parameters better preserves long-tail information, which contradicts the often intuitively argued assumption that simply using larger \pting data and more model parameters would improve low-data learning issues such as long-tail generalization.

\textbf{Contributions:} With CLESS, we propose a \pting data and compute efficient self-supervised contrastive learning objective. Its self-supervision \pting objective head is directly transferable to any end task. This makes CLESS inherently capable of \emph{self-supervised} zero-shot learning, unlike methods which rely on supervised \pting for zero-shot prediction \cite{MTLE, GILE,BERT_Matching_network} or self-supervision approaches which are not capable of zero-shot prediction like RoBERTa or \cite{SIMCLR,CNN_LARGE_SCALE_SSL_PRETRAIN} -- details in \cref{sec:zero-shot}. 
Like T5 \cite{T5}, CLESS models arbitrary NLP tasks as `text-to-text' prediction, but extends on T5 via (a) data efficient contrastive self-supervision and (b) by performing `text-to-text' prediction in embedding space, rather than in token space. The result is an efficient and scalable, yet simple noise contrastive (pre-)training objective, which compared to softmax objectives, as used in e.g. T5, reduces compute costs and enables using infinite class sets -- see (\cref{sec:model}).
Unlike \cite{X-Bert,BERT_Matching_network,FEW_BERT,CNN_LARGE_SCALE_SSL_PRETRAIN}, CLESS does not require external Web-scale \pting data and effectively pretrains on 3 orders of magnitude smaller data than a RoBERTa baseline approach -- see \cref{sec:few-shot}.

\textbf{Findings:} As can be seen in \cref{fig:overview_results}, the contrastive self-supervised \pting objective results in CLESS being more efficient than a fine-tuned RoBERTa model when it comes to the amount of \pting data and labels used. It further leads to better long-tail generalization performance. Lastly, CLESS \pting plus fine-tuning takes 1/5 the amount of time it takes to fine-tune RoBERTa -- see \cref{fig:overview_results}. To understand the factors behind these improvements, we study the impact of model size and self-supervision signal scale-up during self-supervised zero-shot learning in \cref{sec:zero-shot} and how they subsequently affect the models few-shot and supervised learning ability in \cref{sec:few-shot}. We find that \emph{scaling up \pting self-supervision signal amount on small data can be used in place of large \pting data sizes and models}, while training much faster and generalizing better under challenging learning low-data aspects.


\section{Related Work}
To simultaneously satisfy the four capability requirements of self-supervised \pting, long-tail, zero and few-shot learning, CLESS
extends on three machine learning concepts (see \cref{sec:model} for details). (1) We propose generating (pseudo) label embeddings to \emph{treat the predictions of inputs and outputs as well as self-supervision and supervision as the same task} -- i.e. labels are pretrained (word sequence embeddings) while the \pting task prediction head is reusable (transferable) for any new (self or supervised) task. (2) We propose a novel contrastive self-supervised \pting objective to enable \emph{efficient \pting on small datasets and to enhance zero-shot, few-shot and long-tail generalization performance}. (3) CLESS uses a noise contrastive estimation \cite{NCE} objective to enable self-supervised \pting (or fine-tuning) and allow one to \emph{easily vary the amount of self-supervised learning signal}. Thus, instead of adding more data, one can increase self-supervision learning signals instead. Notably, CLESS \pting does not require special learning rate schedules, residuals, normalization, warm-ups or a modified optimizer as *BERT* variations do \cite{BERT, ROBERTA, transformerHardToTrain}. 

\textbf{Self and supervised zero-shot capabilities from \pting:} Large-scale methods like \citet{CNN_LARGE_SCALE_SSL_PRETRAIN, SIMCLR} use large `task data external', self-supervised \pting, but are not zero-shot capable, since they require a newly initialized prediction head per end-task. GPT-3 \cite{GPT3}, enables zero-shot prediction without labeled \pting, but \emph{requires massive \pting data and model parameters}. Mapping labels into embedding space for learning enables zero-shot leaning, however \citet{MTLE, GILE, VisionCP,BERT_Matching_network,augenstein-etal-2018-multi,augenstein-etal-2019-multifc} only pretrain supervised label embedding objectives, which limits learning to only \light{supervised zero-shot transfer} -- i.e.\ no zero-shot learning without labels. 

\textbf{Contrastive learning, mutual information maximization and noise contrastive estimation:} Contrastive objectives can be interpreted as maximizing the lower bound of mutual information between different views of the same data samples \cite{Oord, Contrastive_as_MInfo_max}. These methods often use a form of noise contrastive estimation \cite{NCE_unnormalized}, which replaces softmax normalization by instead learning to contrast positive from negative text instances. A more extensive survey and explanation of contrastive learning in NLP is provided by \citet{rethmeier2021primer}.

\textbf{Long-tail and countering minority bias:} \citet{XCM_CNN} use a label embedding CNN for better many class prediction without pretraining. \citet{X-Bert} combine two Web-scale, externally pretrained models with label embedding matching to further push many-class performance. However, they find that the BERT transformer token-embeddings can not be used as labels embeddings, so they use ELMO word embeddings. Additionally, \citet{hookerCompressedLongtail, COMPRESSION_KILLED_THE_LONG_TAIL} find that the ability to encode long-tail (minority) information is a key ingredient in algorithmic counter minority bias reduction because: (a) ``minority fairness considerations coincide with long-tail information'', while (b) ``compressed models, loose long-tail, minority class prediction quality first'', where (c) this loss is ``often masked by common evaluation methodology and measures''.  
\begin{figure*}[ht]
    \centering
    \includegraphics[width=.98\textwidth, trim=0 0 0 0, clip]{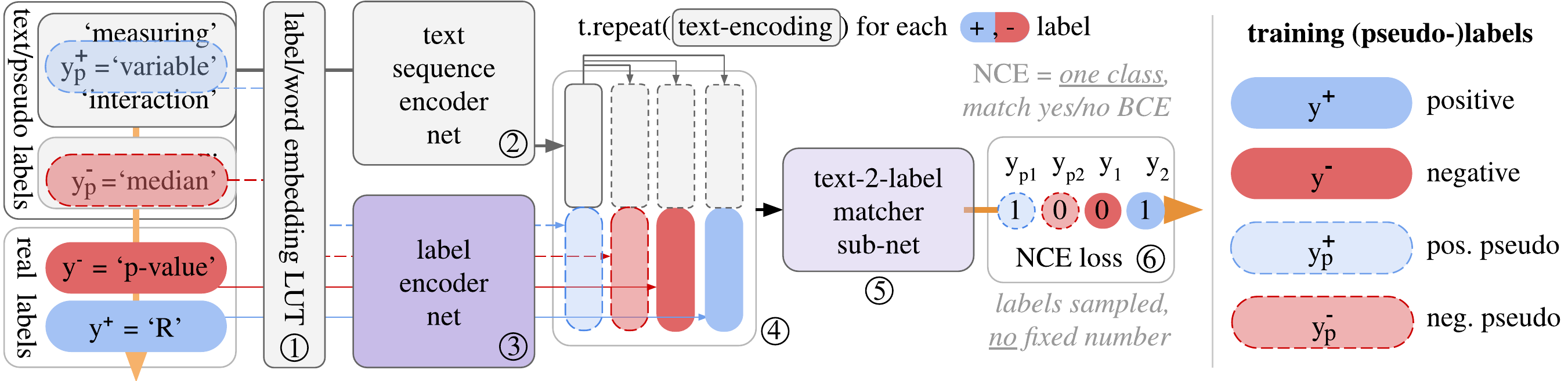}  \caption{\textbf{Contrastive text-sequence-embedding to label-embedding matcher model:} A text (`measuring variable interaction'), and positive (`variable', R) or negative (pseudo) labels (`p-value', `median') are encoded by the word embedding layer $E$ \Circled{1}, where labels are word IDs for lookup. The text embeddings are encoded by a sequence encoder $T$ \Circled{2}, while a label encoder $L$ \Circled{3} encodes $c$ labels. Each text has multiple (pseudo) labels, so the text encoding $\vt_i$ is repeated for, and concatenated with, each label encoding $\vl^\circ_{i,l}$. The resulting batch of `text-embedding, label-embedding' pairs $[[\vt_{i},\vl^\circ_{i,1}], \dots, [\vt_{i},\vl^\circ_{i,c}]]$ \Circled{4} is fed into a `matcher' classifier \Circled{5} that trains a binary cross entropy loss \Circled{6} on multiple label (mis-)matches $\{0,1\}$ per text instance $\vt_{i}$; resulting in a noise contrastive estimation objective (NCE). Text words like `variable' provide self-supervision pseudo labels. Positive and negative (pseudo) labels are sampled from their own or other instance texts in a mini-batch. Unlike \cite{LE_CNN} we use negative sampling and self-supervision as well as a CNN for \Circled{2}.}\label{fig:model}
\end{figure*}

\textbf{Data and parameter efficiency:} CNNs have been found to be more data efficient than Transformers, i.e. (pre)train efficiently from less data, in works by \citet{BlunsomTransformerDataInefficient, BLUNSOM, EFFICARE, CNN_LARGE_SCALE_SSL_PRETRAIN}, while \citet{CLIP} find that replacing a Transformer language encoder with a CNN backbone increased zero-shot data efficiency 3 fold, which is further increased via \emph{supervised} contrastive learning.

\textbf{Contributions:} CLESS proposes a novel, easy to train and adapt self-supervised text encoder model (for small data) by combining noise contrastive estimation with self-supervised, data efficient \pting.
%

\section{CLESS: unified self- and supervision via contrastive `dense-to-dense' text prediction}\label{sec:model}
Below, we describe CLESS (\cref{fig:model}), which learns to contrast labels in embedding space for contrastive self-supervised \pting in NLP. Previous methods that embed labels only use \emph{supervised learning (pre-)training} \cite{MTLE,GILE}, while current self-supervised \pting in NLP focuses on large-scale `task external \pting data' and progressively larger models models \cite{GPT3,BIG_MODELS}. Instead, we propose a \emph{contrastive self-supervision} objective that scales up the \pting self-supervision signal rather than use more data to enable text data efficient \pting of `text-to-text' tasks. While in this work we focus on evaluating CLESS for long-tail prediction, other NLP tasks like question answering or entailment could easily be modeled as $X{=}$ seq1 [sep] seq2, $Y{=}$ `is answer'/`negative entailment'/`neutral'.   

Most NLP models translate back and forth between discrete words and continuous token embeddings by using a softmax computation that is limited to predicting classes known at training time \cite{T5}. However, we require that CLESS in able to zero-shot predict unknown classes, without prior exposure to \emph{any} real labels. This means that after purely \emph{self-supervised \pting}, absent real labels, our model prediction head is pretrained and reusable, or `zero-shot transferable', to arbitrary downstream tasks. This removes a limitation of current \pting methods, which require initialization of one new prediction head per new task \cite{BERT,SIMCLR,CNN_LARGE_SCALE_SSL_PRETRAIN,ROBERTA}. Below, we describe how \emph{CLESS learns to contrast embeddings of real and pseudo labels to unify self- and supervised learning.} 

\textbf{Our first core goal} is to ease small and long-tailed learning by \light{mapping input text words $\vw_i$ and labels $\vw^\circ_{i}$ into the same word representation space, i.e.\ draw them from a shared embedding look-up table $E$ \Circled{1}. We thus replace dense to sparse translations with `embedding-to-embedding' matching \Circled{5}}, to express NLP `text-to-text' tasks \cite{T5} as `dense(text)-to-dense(text)' tasks. Enabling zero-shot predictions requires learning text-to-label embedding pairing \Circled{4} matches \Circled{5}. Hence, the labels $\vw^\circ_{i}$ of a text instance $i$ are replaced by their corpus-internally pretrained fastText or otherwise initialized label(-word) embeddings $\mL^\circ_i$ 
via a label-encoder $L$. Composite label embeddings, like `p-value', are constructed from the mean of the pretrained word embeddings for the words `p' and `value'. During (pre-)training, the label embeddings can be refined to fit a self- or supervised task, since their word embeddings are used within multiple input text instances. 

This automates the embedding of labels, which removes the need to ``manually define label embeddings from word embedding averages over words that describe the label'' as done in previous methods \cite{MTLE,GILE}. Though CLESS supports using manually curated label description embeddings, we instead automatically construct label embeddings to enable self-supervision and scalability. To facilitate contrastive, self-supervised \pting, CLESS treats input words as positive or negative `pseudo labels'. \emph{Positive pseudo labels are words sampled from the current text instance $i$, while negative pseudo labels are words sampled from other text instances $j\neq i$ in a batch}. The pseudo label embeddings are used for noise contrastive estimation \cite{NCE}, resulting in a contrastive, partial autoencoding objective. This has two implicit advantages. One, \emph{labels are pretrained, because their comprising word embeddings are}. Two, unforeseen new labels can be inferred via methods like \citet{FASTTEXT} or \citet{TIMORW}.

As outlined visually, left to right in \cref{fig:model}, learning multi-label classification then becomes a contrastive learning problem of \emph{matching the word-sequence embedding $\vt_i$} of text $i$ \Circled{2}, with its $c$ label (word-sequence) embeddings $\mL^\circ_{i} = \{\vl^\circ_{i,1}, \ldots \vl^\circ_{i,c}\}$ \Circled{3}, by feeding $c$ text-vs-label combinations $[[\vt_{i},\vl^\circ_{i,1}], \dots, [\vt_{i},\vl^\circ_{i,c}]]$ \Circled{4} to a binary classifier $M$ \Circled{5} for matching. This means that instead of predicting $c$ classes at once, we predict a batch of $c$, single-class, binary classifications using binary cross entropy \Circled{6}. Thus, the number of classes $c$ does not need to be constant across instances $i$.
The details of Steps \Circled{1} to \Circled{6} are as follows.

To train a binary match classifier (\Circled{5},\Circled{6}), we need an input text paired with its accompanying positive and negative labels. 
A text instance $i$ is represented as a vector of word embedding lookup IDs $\vw_i = \{w_{i,1}, \ldots w_{i,s}\}$. For each text instance $i$ we wish to classify, we need $g$ positive label-word IDs $\vw^-_i = \{w_{i,1}^+, \ldots w_{i,g}^+\}\in R^{g}$. Here, a single label such as `p-value' is made up of two lookup IDs for the words `p' and `value' that select word embeddings in the embedding lookup table $E$ \Circled{1}. We also need $b$ negative (bad) label-word IDs $\vw^+_i = \{w_{i,1}^-, \ldots w_{i,b}^-\}\in R^{b}$. Then, for each text instance $i$, we combine the $g$ positive and $b$ negative label-word IDs into a single label embedding lookup vector
\begin{align*}
    \vw^\circ_i = \{\vw_i^+ \oplus \vw_i^-\} \in \R^{c=g+b} 
\end{align*}
To indicate ground truth positive and negative labels for later match classification \Circled{6}, we also need a $g$-sized vector of ones $\vone\in \R^g$ and a $b$-sized zero vector $\vzero\in \R^b$, concatenated into a class indicator
\begin{align*}
    \mathbb{I}_i = \{\bm{1} \oplus \bm{0}\}\in \N_0^{c=g+b}  \tag{\Circled{6}}   
\end{align*}
Together, $\vw^\circ_i \in \R^{c=g+b} $ and $\mathbb{I}_i \in \N_0^{c=g+b}$ describe the label embeddings and binary labels for a single text instance $i$.
Both the text (word) indices $\vw_i$ and the label embedding lookup IDs $\vw^\circ_i$ are passed through a shared `word-or-label embedding' lookup table $E$ \Circled{1}.
\begin{align*}
    \text{label-word/ word embeddings} := &\ E(\vw^\circ_i), E(\vw_i) \tag{\Circled{1}}\\
\end{align*}
Then the embeddings are passed through the text-sequence encoder $T$ and the label encoder $L$. 
Per text instance $i$, the text-encoder $T$ produces a (single) text embedding vector
\begin{align*}
    \vt_i = T(E(\vw_i)) \tag{\Circled{2}}
\end{align*}
The label-encoder $L$ produces $c=g+b$ label embedding vectors $\vl_{i,\cdot}^\circ \in \R^{c}$ that form a label embedding matrix:
\begin{align*}
    \mL^\circ_i = [\vl_{i,1}^+, \ldots , \vl_{i,g}^+, \vl_{i,1}^-, \ldots , \vl_{i,b}^-] \leftarrow L(E(\vw^\circ_i)) \tag{\Circled{3}}
\end{align*}
As text-encoder we use a (CNN$\rightarrow$max-k-pooling$\rightarrow$ ReLU) sub-network, while the label-encoder is simply an (average-pool) operation, since a single label $\vl_{i,\cdot}^\circ$, e.g.\ `multi'-`label', can consist of multiple word IDs $\vw^\circ_{i,\cdot}$.
To compare how similar the text-embedding $\vt_i$ is to each label embedding $\vl^\circ_{i,\cdot}$, we repeat $\vt_i$ $c$ times and concatenate text and label embeddings to get a text-vs-label-embedding matrix 
\begin{align*}
    \mM_i = [[\vt_i,\vl^+_{i,1}], \ldots, [\vt_i,\vl^-_{i,c}]] \tag{\Circled{4}}
\end{align*}
This text-label paring matrix $\mM_i$ is then passed through the matcher network $M$ \Circled{5} to produce a batch of $c$ probabilities
\begin{equation*}
    \vp_{i} = \{\sigmoid(M(\mM_{i,1})), \ldots, \sigmoid(M(\mM_{i,c}))\} \tag{\Circled{5}}    
\end{equation*}
We use binary cross entropy (BCE) as a loss between $c=b+g$ predicted label probabilities $\vp_i$ and label indicators $\mathbb{I}_i$
\begin{align*}
-\frac{1}{c}\sum_{l=1}^{c} \mathbb{I}_{i,l} \cdot log(\vp_{i,l}) + (1-\mathbb{I}_{i,l}) \cdot log(1-\vp_{i,l}) \tag{\Circled{6}}
\end{align*}
This BCE over positive and negative (pseudo) labels produces a \emph{noise contrastive estimation} (NCE) as in \cite{NCE_unnormalized}, but we use $g$ positives instead of the standard $1$ positive.
NCE replaces the computationally expensive normalization of the softmax \cite{NCE_unnormalized} by instead learning a (mis-)match between (text-embedding, label-embedding) pairs. This undersamples the negative labels in training, to save compute and enable using infinite class (output) sets. For supervised fine-tuning on real labels, we use all positive labels, but undersample negative classes. For self-supervised \pting, we sample $g=b$ input batch words as pseudo labels. Here positives are samples from the current text instance, i.e. sampled from a block of similar data as suggested in \citet{ContrastiveLearningLimitations} to improve contrastive learning. Negative pseudo labels are sampled words that appear in other texts from a mini-batch, but do not appear in the current instance text.

\textbf{Our second core goal:} Once inputs $X$ and outputs $Y$ are pretrained (well initialized), \emph{training via pseudo label self-supervision enables one to also pretrain the encoder model $\Theta=\{\Circled{2} \ldots \Circled{5}\}$ and the prediction (matcher) module \Circled{6}}. As a result, the entire model becomes pretrained (well initialized) allowing one to handle self-supervised, supervised, few and zero-shot learning in a unified manner.
%
\begin{figure}[t]
    \centering
    \includegraphics[width=1\linewidth, trim=6 6 6 6, clip]{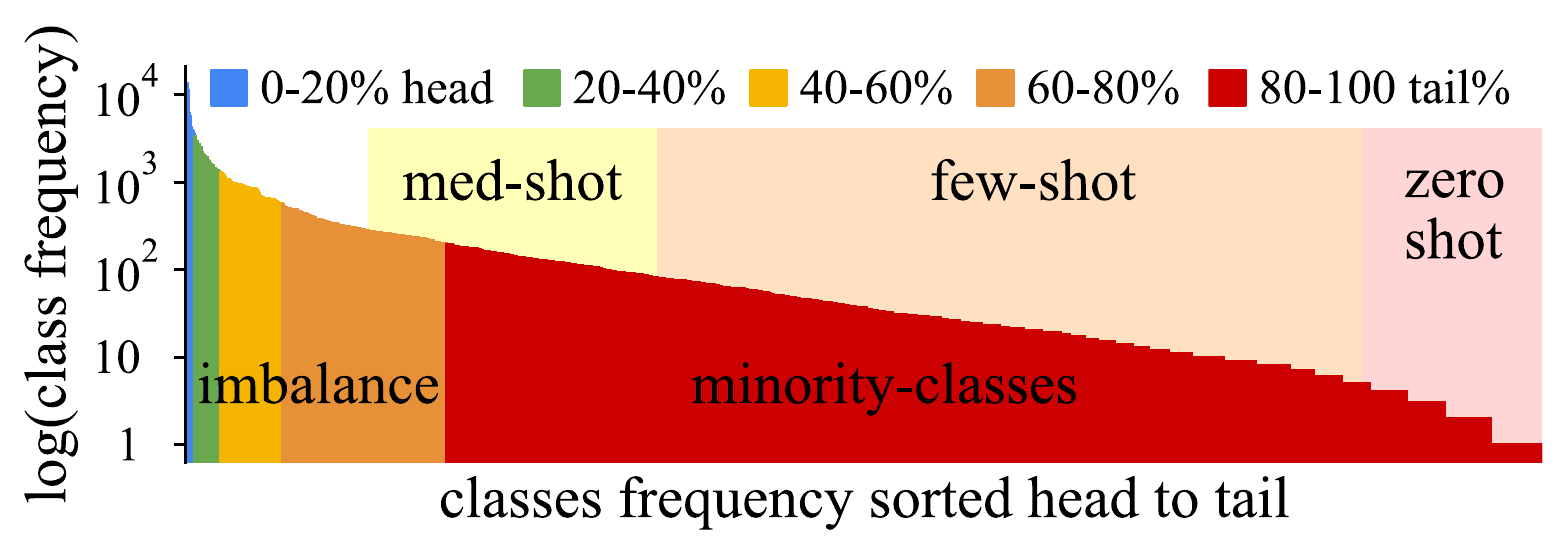}
    \caption{\textbf{Head to long-tail as 5 balanced class bins:} We bin classes by label frequency. Each bin contains equally many active label occurrences. Classes within a bin are imbalanced and become few-shot or zero-shot towards the tail, especially after train/dev/test splitting. Class frequencies in log scale -- task data details in \cref{sec:data}.}\label{fig:longtail_dist}
\end{figure}
\section{Small, long-tailed `text-to-text' (and label) prediction as a challenging test bed}\label{sec:data}
To research input data, label and long-tail learning efficiency for `text-to-text' \pting models we choose a small multi-label question tag prediction dataset as a test bed. We use the ``Questions from Cross Validated''\footnote{https://www.kaggle.com/stackoverflow/statsquestions} dataset, where machine learning concepts are tagged per question. This dataset has three appropriate qualities: \emph{it is small-scale, long-tailed, and entails solving a challenging, noisy `text-to-text' prediction task}. As with many real world problems, labels are vague, since tagging was crowd-sourced. This means that determining the correct amount of tags per question (label density) is hard, even for humans.
This task currently has no published state-of-the-art. As seen in \cref{fig:longtail_dist}, the datasets' class occurrence frequencies are highly long-tailed, i.e.\ the 20\% most frequently occurring classes result in 7 `head' classes, while the 20\% least frequent (rightmost) label occurrences cover 80\% or 1061/1315 of classes. Tags are highly sparse -- at most 4 out of 1315 tags are labeled per question. Word embeddings are pretrained with fastText -- details in supplement. The dataset has 85k questions with 244k positive labels.
\begin{figure*}[t]
\centering
\begin{minipage}{1\textwidth}
    \includegraphics[width=.49\linewidth, trim=0 0 0 0, clip]{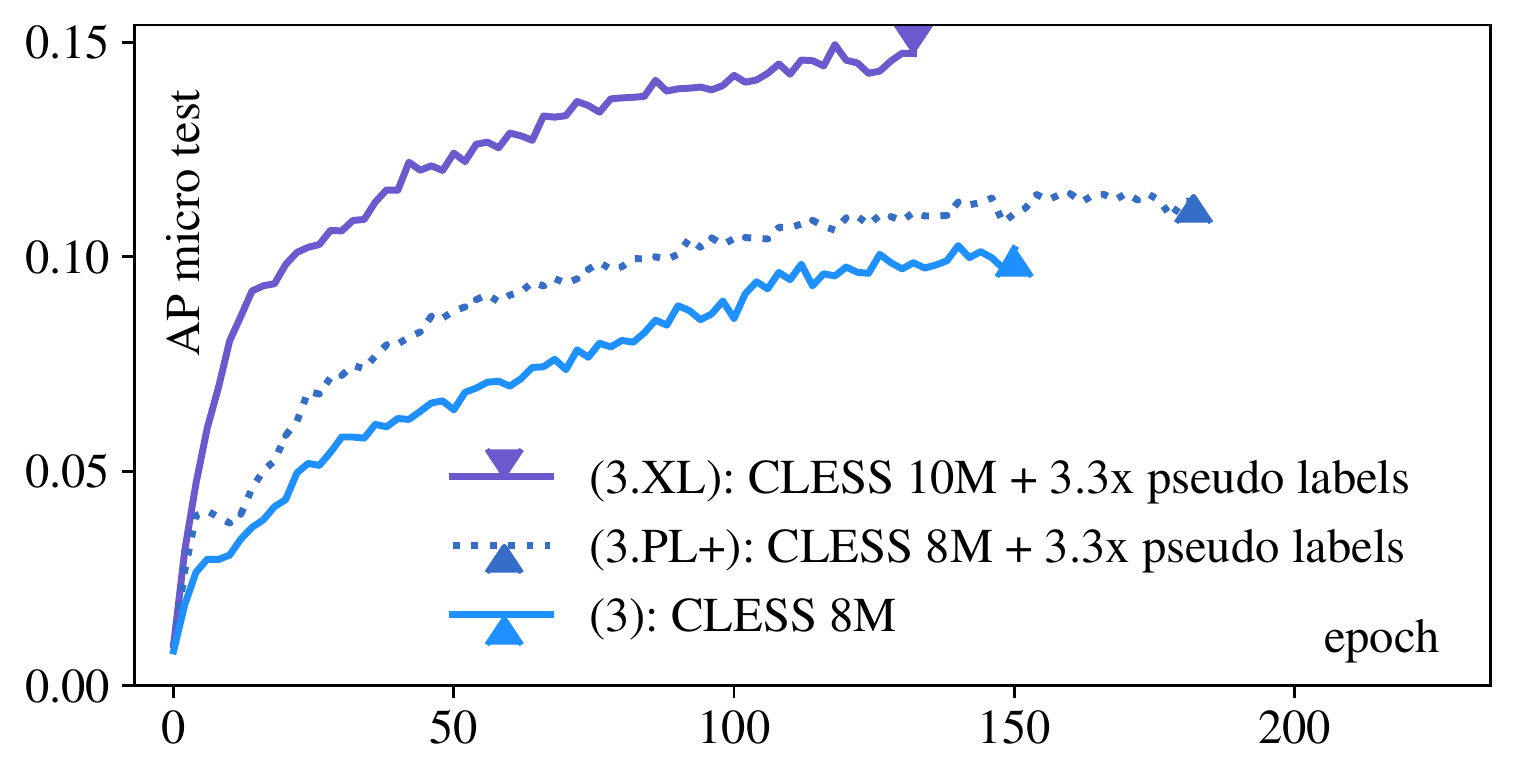}\hspace{4.2mm}\includegraphics[width=.49\linewidth, trim=0 0 0 0, clip]{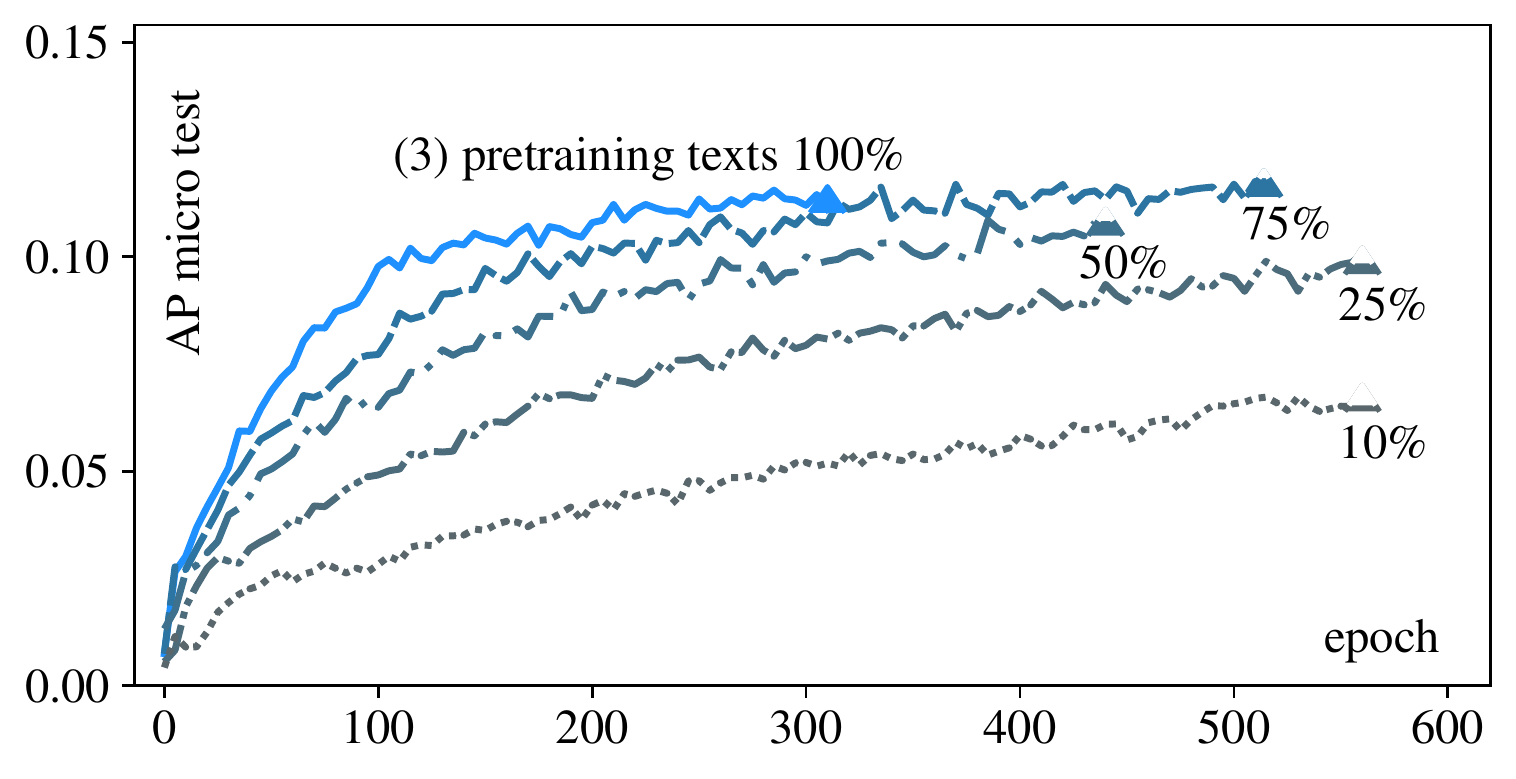}   
    \caption{\textbf{Zero-shot performance by model size, pseudo label amount and \pting data amount (X-efficiency):} \textbf{Left:} The zero-shot performance of the self-supervised \pting base model (3) is increased when, adding more self-supervision pseudo labels (3.PL+) and widening the network (3.XL). \textbf{Right:} When only using only a proportion of the \pting input data texts to pretrain model (3), its zero-shot learning is slowed down, but \emph{still converges towards the 100\% for all but the most extreme \pting data  reductions}.}\label{fig:few-zero_shot}
\end{minipage}
\end{figure*}

\section{Experimental setup and metrics}
Our primary research \emph{goal with this work is to better understand the relations between \pting data size, self-supervision signal amount and model size} in comparison to current large-scale \pting methods. Here we describe Average Precision as a scalable evaluation metric. In the next sections analyze (a) \pting data ($X$-)efficiency in a zero-shot setup, (b) label efficiency as few-shot learning and finally (c) long-tail generalization efficiency.   

%
\textbf{Long-tail evaluation metrics and challenges:}
Long-tail, multi-label classification is challenging to evaluate because (i) top-k quality measures 
neglect performance loss on long-tailed minority classes as \citet{COMPRESSION_KILLED_THE_LONG_TAIL,MITIGATING_BIAS} point out.
Furthermore, (ii) measures like $ROC_{AUC}$ overestimate performance under class imbalance \cite{APIMBALANCES, APIMBALANCES2} and (iii) discrete measures like F-score are not scalable, as they require discretization threshold search under class imbalance.
Fortunately, the Average Precision score $AP=\sum_{n} (R_n - R_{n-1})P_n$ addresses issues (i-iii), where $P_n$ and $R_n$ are precision and recall at the $n$th threshold.
As weighting we choose $AP_{micro}$ and $AP_{macro}$ as they are the lowest scoring, hardest to optimize variants, which helps to reduce optimistic evaluation. 
%
%

\section{Results}
In this section, we analyze the \pting data ($X-$)efficiency of CLESS under a zero-shot learning setup to study the relation between model size and self-supervision signal amount (\cref{sec:zero-shot}. Then we compare RoBERTa and CLESS variants in a few-shot and fully supervised setting to study their label ($Y-$)efficiency. Finally, we study the long-tail efficiency of CLESS variants and RoBERTa. To do so, we split the dataset into 80/10/10 for training, development and test set. \emph{Test scores or curves are reported for models that have the best development set average precision score $AP_{micro}$ over all 1315 classes.} RoBERTa uses 125 million parameters and pretrained on 160GB of external text data. CLESS models use 8-10 million parameters and pretrained on 60MB of text data. As is standard for establishing guessing performance under class imbalance, we use a ZeroR classifier, which predicts the majority label per class. The ZeroR $AP_{micro}$ and $AP_{macro}$ on this dataset are $0.002$ since a maximum of 4 in 1315 classes are active per instance -- i.e.\ this low guessing performance underlines the challenge of this learning task.

\subsection{Zero-shot: more contrastive \pting boosts data efficiency.}
\label{sec:zero-shot}
Here, we study the self-supervised \pting data efficiency ($X$-efficiency) of CLESS -- i.e.\ its zero-shot performance on 6-60MB of \pting data\footnote{Since the goal of this work is efficient learning, we do not re-pretrain a RoBERTa for zero-shot prediction. Firstly, because this retraining would cost 500 times the training time of CLESS, with only 100 positive plus negative pseudo labels. Secondly, because RoBERTa uses external \pting data, which is not \pting data efficient, \emph{and thus is not in line with the goal of this work.}}.
To do so, we pretrain CLESS variants via pseudo labels and evaluate their zero-shot performance on real test set labels from \cref{sec:data}.
In \cref{fig:few-zero_shot} we analyze the effect of scaling the pseudo label self-supervision amount and its relation to the number of model parameters. We see that the CLESS 8M model (3), pretrained with 8 million parameters and 150 pseudo labels, achieves around $.1 AP_{micro}$ on the test real labels as zero-shot performance. 
When \emph{increasing the number of self-supervised word pseudo labels} from 150 to 500 in model (3.PL+), \emph{the model gains zero-shot performance} (middle curve), without requiring more parameters. When raising model parameters to 10M in (3.XL), the zero-shot performance increases substantially (top curve). Thus, for self-supervised zero-shot performance, both increased self-supervision signal amount and model size matter, whereas previous research only linked zero-shot performance to largely increased model sizes in Web-scale \pting such as GPT-3 \cite{GPT3}.

\textbf{When zero-shot leaning with less data, pretrain longer!} Further, in the \emph{right plot} of \cref{fig:few-zero_shot} we see that when \pting a model (3) on only portions ($100\%, \ldots ,10\%$) of \pting text inputs $X$, i.e.\ in an increasingly low-data zero-shot setup, the model still converges towards the original 100\% performance.
As expected, convergence slows with smaller \pting text data portions since each data reduction also implies seeing fewer pseudo label signals per epoch. Thus, in the right-side plot, we allow for $5x$ more waiting epochs for early stopping than in the left plot. \light{This provides a promising insight for self-supervised \pting on small data, which, if designed appropriately, can favorably initialize models for zero-shot learning from very small text sizes and without real label \pting}, as was previously required in \citet{GILE,VisionCP}.

\subsection{Few-shot: contrastive \pting improves label efficiency}\label{sec:few-shot}
Here we study how \emph{`task-internal'} pseudo label \pting of CLESS compares against not \pting CLESS and against a large-scale, `task-external' data \pting model like RoBERTa. For the few-shot setup, we use 100\%, 50\% and 10\% of labels for supervised learning. This means that if labels were common in the 100\% setup, they now become increasingly rare or few-shot in the 10\% setup, since the smaller set is long-tail distributed as seen in \cref{fig:longtail_dist}.

\begin{figure}[t]
    \includegraphics[width=\linewidth, trim=6 7 6 6, clip]{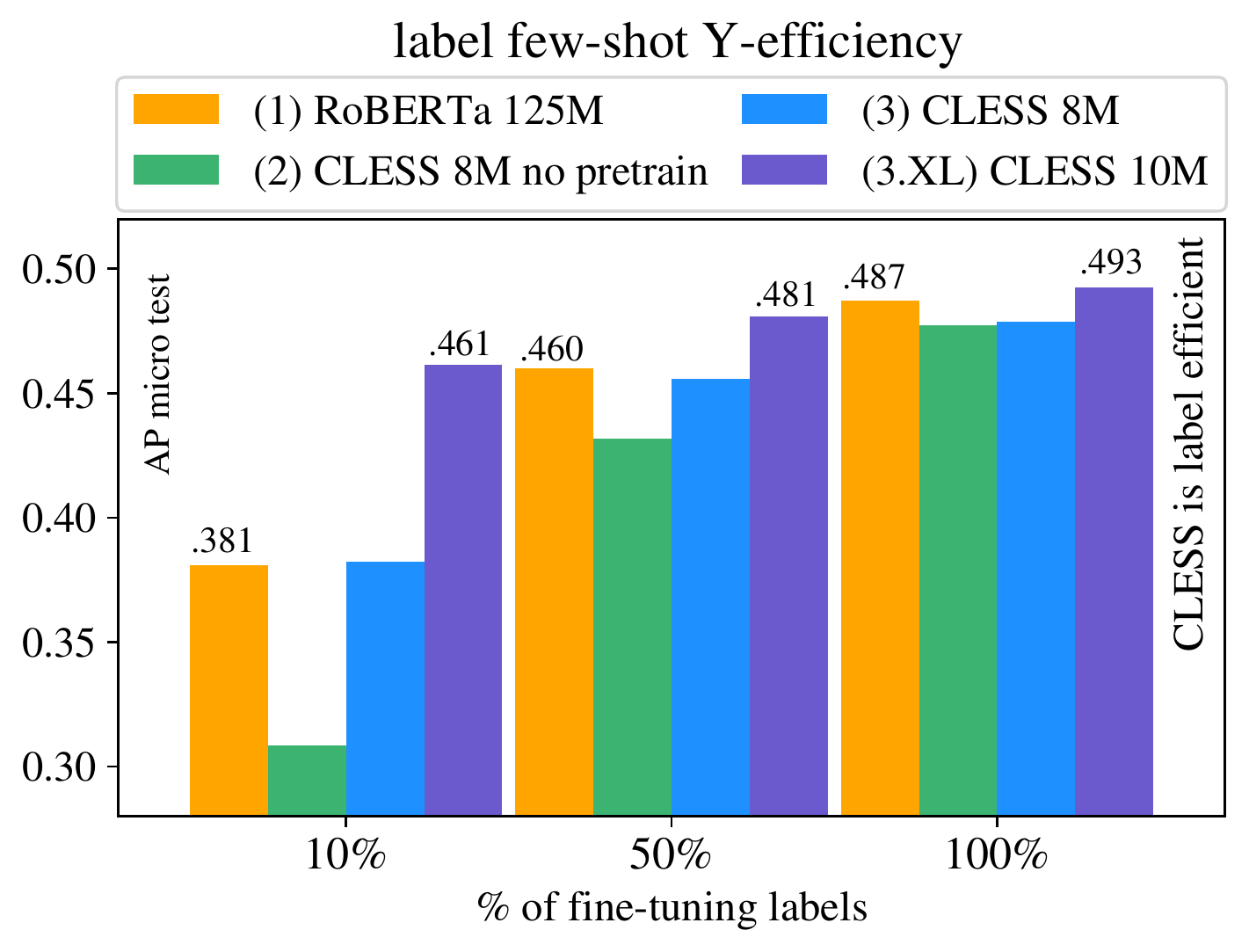}\hspace{4.2mm}
    \caption{\textbf{Few-shot efficiency:} (1) externally pretrained RoBERTa. (2) CLESS without \pting. (3) CLESS with \pting and parameter otherwise like (2). (3.XL) CLESS with more parameters as in (\cref{sec:zero-shot}). $AP_{micro\_test}$ scores for few-shot portions: 100\%, 50\%, 10\% of training samples with real labels. CLESS 10M outperforms RoBERTa, with less \pting data, parameters and total training time -- 25h CLESS vs.\ 126h RoBERTa (see \cref{fig:overview_results}).}\label{fig:Few_shot}
\end{figure}

In \cref{fig:Few_shot}, we see that when using full supervision (100\%) all models perform similarly, with CLESS (3.XL) slightly outperforming RoBERTa (.493 vs.\ .487) $AP_{micro\_test}$.
For few-shot learning (10\%, 50\%) we see that CLESS 3.XL retrains $.461/.493{=}0.935\%$ of its original performance when using only 10\% of labels for fine-tuning, while RoBERTa and CLESS 8M both retain around 77\%. This demonstrates that a larger model \pting size and self-supervision signal amount not only boosts zero-shot learning performance as in \cref{fig:few-zero_shot}, but also translates into much better few-shot performance.
Noticeably, the only non-pretrained model (2), performs much worse than the rest in the more restricted few-shot scenarios.
Since models (2) and (3) use the same hyperparameters and only differ in being pretrained (3) or not being pretrained (2), this demonstrates that `task data internal' \pting is important for label efficient learning.

\subsection{Tail generalization: is improved by contrastive \pting}\label{sec:longtail}
Here, we study the effect of contrastive self-supervised \pting on long-tail generalization. Plotting individual scores for 1315 classes is unreadable. Instead, we sort classes from frequent to rare and assign them to one of five $20\%$ class frequency bins, such that all bins contain the same amount of positive real labels (label occurrences). As seen in \cref{fig:longtail_dist}, this means that the head bin (left) contains the most frequent $7/1315{=}0.5\%$ classes, while the tail contains the most rare $1061/1315{=}80.7\%$ classes. That is, the tail contains increasingly more minority information classes. We balance inter-bin label frequency to make the five bins directly comparable.
\begin{figure}[t]
    \centering
    \includegraphics[width=1\linewidth, trim=6 7 6 6, clip]{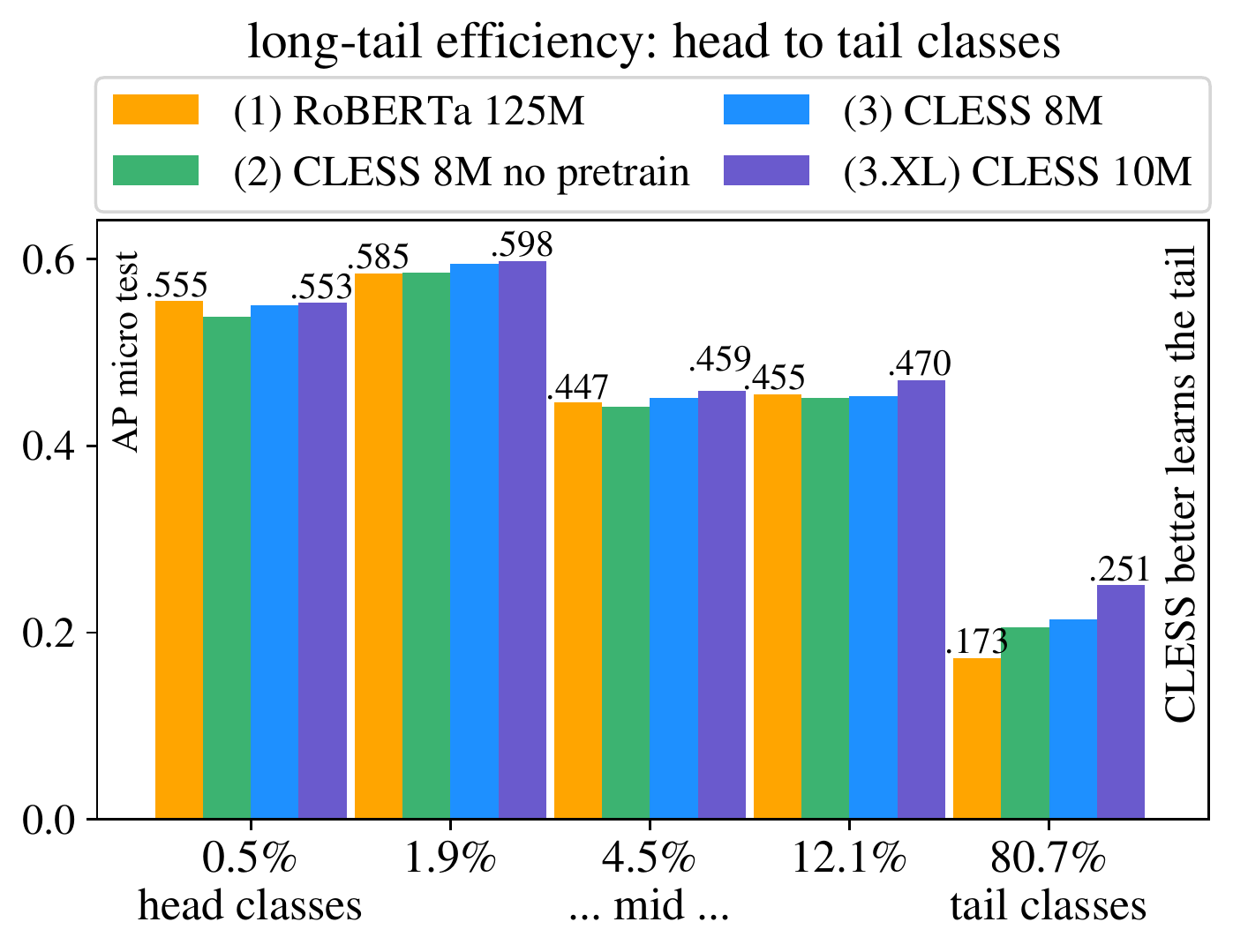}
    \caption{\textbf{Long-tail performance of RoBERTa, non-pretrained (2) and pretrained (3) and (3.XL) CLESS:} Over all five head to tail class bins -- see \cref{fig:longtail_dist} -- the non-pretrained CLESS (2), performs worst. The tail class bin contains 80.7\% or 1062/1315 classes. The largest pretrained CLESS model (3.XL) outperforms RoBERTa in tail and mid class prediction, while performing almost on par for the $7/1315{=}0.5\%$ (most common) head classes.}\label{fig:longtail}
\end{figure}

\textbf{CLESS contrastive \pting improves performance over RoBERTa for 99.5\% of classes:} In \cref{fig:longtail} we see how RoBERTa and CLESS perform over the class head-to-tail bins defined in \cref{fig:longtail_dist}. The 125M parameter RoBERTa model performs best on the most common $7/1315{=}0.5\%$ classes by a small margin. The largest pretrained CLESS model (3.XL) with 10M parameters performs best on all but the 7 head classes -- i.e.\ it outperforms RoBERTa on 99.5\% of classes. CLESS (3.XL) performs especially well on the $1062/1315{=}80.7\%$ of tail classes that contain the 20\% rarest label occurrences. From the existing literature on large data pretrained models, our intuitive  assumption would be that large \pting data implicitly makes a model like RoBERTa learn more long-tail information than is learnable from small \pting data. However, the results dispel this intuition, since CLESS produces a .251 average class bin $AP_{micro\_test}$ performance compared to .173 for RoBERTa. This, qualifies three insights. One and two, that neither using large data \pting nor using magnitudes more model parameters (e.g.\ 10M vs 125M) can guarantee long-tail information capture. Third, we instead demonstrated that \pting via contrastive self-supervision can efficiently learn long-tail generalization, despite using very limited \pting data.

We also see that CLESS 8M outperforms RoBERTa on the tail, but is itself worse than CLESS 10M, which if we recall its \pting in \cref{sec:zero-shot}, uses more parameters and 3.3x more self-supervision pseudo labels for \pting. This allows us to gather another insight. Four, that contrastive self-supervised \pting model size and signal amount correlate with improved long-tail generalization performance. Put another way, for algorithmic bias caused by failure to preserve long-tail information, as described in \cite{hookerCompressedLongtail,COMPRESSION_KILLED_THE_LONG_TAIL,HookerPatterns}, CLESS can samples more self-supervision pseudo labels to guarantee generating a more rare self-supervised learning signals.

\section{Conclusion}
CLESS demonstrates that a small contrastive self-supervision model can efficiently learn challenging tasks like zero-shot, few-shot and long-tail prediction, without requiring large \pting dataset, models or training times.
Works like GPT-3 \cite{GPT3} have connected zero-shot and few-shot learning improvements to \pting large models on large-scale `task-external' \pting dataset. We study such improvement can be produced more data efficiently. In \cref{sec:zero-shot} we instead increase self-supervision signal amount during \pting to gain better zero-shot performance. Increasing model size by 25\% further raises CLESSs' zero-shot performance by 49\%. In \cref{sec:few-shot} we demonstrate that the increase in \pting signal and model size also translate into better few-shot performance, where a CLESS pretrained on 60MB of `task-internal' data markedly outperforms a RoBERTa pretrained on 160GB of `task-external' data. Previous works like \cite{GPT3,FEW_BERT,TIMO} has shown that large data \pting improves few-shot learning. CLESS demonstrates that a simple contrastive learning objective can produce similar or better results, while only using the task data at hand and being many times faster to compute -- see \cref{fig:overview_results}. Finally, we show that contrastive self-supervised \pting enables a small model to efficiently learn better long-tail generalization. These results demonstrates that increased self-supervision can replace increased \pting data and model size, especially when `task external' \pting data is undesirable.
In future extensions we envision to apply CLESS to (pre-)training of low-data applications, e.g. in medicine \cite{EFFICARE}, or where new labels emerge at test time; as is the case in hashtag prediction \cite{conf/cikm/MaSYC14}. 